\documentclass[10pt,twocolumn,letterpaper]{article}

\usepackage{cvpr}
\usepackage{times}
\usepackage{epsfig}
\usepackage{graphicx}
\usepackage{amsmath}
\usepackage{amssymb}
\usepackage{subfigure}
\usepackage{multirow}
\usepackage[english]{babel}
\usepackage{url}

\usepackage[breaklinks=true,bookmarks=false]{hyperref}

\cvprfinalcopy 

\def\cvprPaperID{****} 

\ifcvprfinal\fi
\begin{document}

\title{Learning Spatio-Temporal Representation with Local and Global Diffusion\thanks{{\small This work was performed at JD AI Research.}}}

\author{Zhaofan Qiu$^{\dagger}$, Ting Yao$^{\ddagger}$, Chong-Wah Ngo$^{\S}$, Xinmei Tian$^{\dagger}$, and Tao Mei$^{\ddagger}$\\
\parbox{20em}{\small\centering $^{\dagger}$ University of Science and Technology of China, Hefei, China}\\
\parbox{40em}{\small\centering $^{\ddagger}$ JD AI Research, Beijing, China~~~~~~~~~~~~~~~~~~~$^{\S}$ City University of Hong Kong, Kowloon, Hong Kong}\\
{\tt\small \{zhaofanqiu, tingyao.ustc\}@gmail.com, cscwngo@cityu.edu.hk, xinmei@ustc.edu.cn, tmei@live.com}
}

\maketitle
\thispagestyle{empty}

\begin{abstract}
Convolutional Neural Networks (CNN) have been regarded as a powerful class of models for visual recognition problems. Nevertheless, the convolutional filters in these networks are local operations while ignoring the large-range dependency. Such drawback becomes even worse particularly for video recognition, since video is an information-intensive media with complex temporal variations. In this paper, we present a novel framework to boost the spatio-temporal representation learning by Local and Global Diffusion (LGD). Specifically, we construct a novel neural network architecture that learns the local and global representations in parallel. The architecture is composed of LGD blocks, where each block updates local and global features by modeling the diffusions between these two representations. Diffusions effectively interact two aspects of information, i.e., localized and holistic, for more powerful way of representation learning. Furthermore, a kernelized classifier is introduced to combine the representations from two aspects for video recognition. Our LGD networks achieve clear improvements on the large-scale Kinetics-400 and Kinetics-600 video classification datasets against the best competitors by 3.5\% and 0.7\%. We further examine the generalization of both the global and local representations produced by our pre-trained LGD networks on four different benchmarks for video action recognition and spatio-temporal action detection tasks. Superior performances over several state-of-the-art techniques on these benchmarks are reported. Code is available at: \url{https://github.com/ZhaofanQiu/local-and-global-diffusion-networks}.
\end{abstract}

\section{Introduction}
Today's digital contents are inherently multimedia. Particularly, with the proliferation of sensor-rich mobile devices, images and videos become media of everyday communication. Therefore, understanding of multimedia content becomes highly demanded, which accelerates the development of various techniques in visual annotation. Among them, a fundamental breakthrough underlining the success of these techniques is representation learning. This can be evidenced by the success of Convolutional Neural Networks (CNN), which demonstrates high capability of learning and generalization in visual representation. For example, an ensemble of residual nets \cite{he2015deep} achieves 3.57\% top-5 error on ImageNet test set, which is even lower than 5.1\% of the reported human-level performance. Despite these impressive progresses, learning powerful and generic spatio-temporal representation remains challenging, due to larger variations and complexities of video content.

\begin{figure}[!tb]
   \centering {\includegraphics[width=0.45\textwidth]{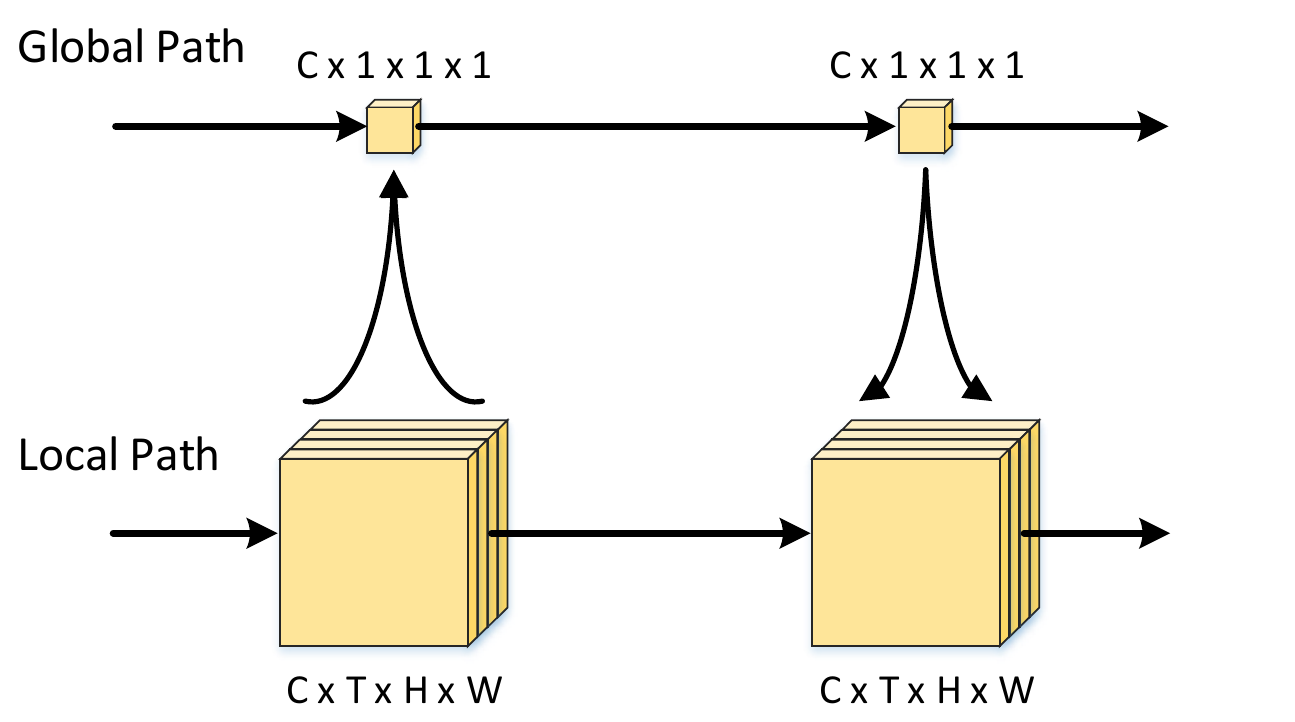}}
   \caption{\small The schematic illustration of the Local and Global Diffusion block. The diffusion between local and global paths enrich the representation learnt on each path. }
   \label{fig:intro}
   \vspace{-0.25in}
\end{figure}

A natural extension of CNN from image to video domain is by direct exploitation of 2D CNN on video frames \cite{karpathy2014large,simonyan2014two,wang2016temporal} or 3D CNN on video clips \cite{ji20133d,qiu2017learning,qiu2018learning,tran2015learning}. An inherent limitation of this extension, however, is that each convolution operation, either 2D or 3D, processes only a local window of neighboring pixels. As window size is normally set to a small value, the holistic view of field cannot be adequately captured. This problem is engineered by performing repeated convolution and pooling operations to capture long-range visual dependencies. In this way, receptive fields can be increased through progressive propagation of signal responses over local operations. When a network is deep, the repeated operations, however, post difficulty to parameter optimization. Concretely, the connection between two distant pixels are only established after a large number of local operations, resulting in vanishing gradient.

In this paper, we present Local and Global Diffusion (LGD) networks -- a novel architecture to learn spatio-temporal representations capturing large-range dependencies, as shown in Figure \ref{fig:intro}. In LGD networks, the feature maps are divided into local and global paths, respectively describing local variation and holistic appearance at each spatio-temporal location. The networks are composed of several staked LGD blocks of each couples with mutually inferring local and global paths. Specifically, the inference takes place by attaching the residual value of global path to the output of local feature map, while the feature of global path is produced by linear embedding of itself with the global average pooling of local feature map. The diffusion is constructed at every level from bottom to top such that the learnt representations encapsulate a holistic view of content evolution. Furthermore, the final representations from both paths are combined by a novel kernel-based classifier proposed in this paper.

The main contribution of this work is the proposal of the Local and Global Diffusion networks, which is a two-path network aiming to model local and global video information. The diffusion between two paths enables the capturing of large-range dependency by the learnt video representations economically and effectively. Through an extensive set of experiments, we demonstrate that our LGD network outperforms several state-of-the-art models on six benchmarks, including Kinetics-400, Kinetics-600, UCF101, HMDB51 for video action recognition and J-HMDB, UCF101D for spatio-temporal action detection.

\section{Related Work}
We broadly categorize the existing research in video representation learning into hand-crafted and deep learning based methods.

\textbf{Hand-crafted representation} starts by detecting spatio-temporal interest points and then describing them with local representations. Examples of representations include Space-Time Interest Points (STIP) \cite{laptev2005space}, Histogram of Gradient and Histogram of Optical Flow \cite{laptev2008learning}, 3D Histogram of Gradient \cite{klaser2008spatio}, SIFT-3D \cite{scovanner20073} and Extended SURF \cite{willems2008efficient}. These representations are extended from image domain to model temporal variation of 3D volumes. One particularly popular representation is the dense trajectory feature proposed by Wang \emph{et al.}, which densely samples local patches from each frame at different scales and then tracks them in a dense optical flow field \cite{wang2013action}. These hand-crafted descriptors, however, are not optimized and hardly to be generalized across different tasks of video analysis.

The second category is \textbf{deep learning based video representation}. The early works are mostly extended from image representation by applying 2D CNN on video frames. Karparthy \emph{et al.} stack CNN-based frame-level representations in a fixed size of windows and then leverage spatio-temporal convolutions for learning video representation \cite{karpathy2014large}. In \cite{simonyan2014two}, the famous two-stream architecture is devised by applying two 2D CNN architectures separately on visual frames and staked optical flows. This two-stream architecture is further extended by exploiting convolutional fusion \cite{feichtenhofer2016convolutional}, spatio-temporal attention \cite{li2018unified}, temporal segment networks \cite{wang2016temporal,wang2018temporal} and convolutional encoding \cite{diba2017deep,qiu2017deep} for video representation learning. Ng \emph et al. \cite{yue2015beyond} highlight the drawback of performing 2D CNN on video frames, in which long-term dependencies cannot be captured by two-stream network. To overcome this limitation, LSTM-RNN is proposed by \cite{yue2015beyond} to model long-range temporal dynamics in videos. Srivastava \emph{et al.} \cite{srivastava2015unsupervised} further formulate the video representation learning task as an autoencoder model based on the encoder and decoder LSTMs.

The aforementioned approaches are limited by treating video as a sequence of frames and optical flows for representation learning. More concretely, pixel-level temporal evolution across consecutive frames are not explored. The problem is addressed by 3D CNN proposed by Ji \emph{et al.} \cite{ji20133d}, which directly learns spatio-temporal representation from a short video clip. Later in \cite{tran2015learning}, Tran \emph{et al.} devise a widely adopted 3D CNN, namely C3D, for learning video representation over 16-frame video clips in the context of large-scale supervised video dataset. Furthermore, performance of the 3D CNN is further boosted by inflating 2D convolutional kernels \cite{carreira2017quo}, decomposing 3D convolutional kernels \cite{qiu2017learning,tran2018closer} and aggregated residual transformation \cite{hara2018can}.

Despite these progresses, long-range temporal dependency beyond local operation remains not fully exploited, which is the main theme of this paper. The most closely related work to this paper is \cite{wang2018non}, which investigates the non-local mean operation proposed in \cite{buades2005non}. The work captures long-range dependency by iterative utilization of local and non-local operations. Our method is different from \cite{wang2018non} in that local and global representations are learnt simultaneously and the interaction between them encapsulates a holistic view for the local representation. In addition, we combine the final representations from both paths for more accurate prediction.

\section{Local and Global Diffusion}
We start by introducing the Local and Global Diffusion (LGD) blocks for representation learning. LGD is a cell with local and global paths interacting each other. A classifier is proposed to combine local and global representations. With these, two LGD networks, namely LGD-2D and LGD-3D deriving from temporal segment networks \cite{wang2016temporal} and pseudo-3D convolutional networks \cite{qiu2017learning}, respectively, are further detailed.

\subsection{Local and Global Diffusion Blocks}
Unlike the existing methods which stack the local operations to learn spatio-temporal representations, our proposed Local and Global Diffusion (LGD) model additionally integrates the global aspect into video representation learning. Specifically, we propose the novel neural networks that learn the discriminative local representation and global representation in parallel while combining them to synthesize new information. To achieve this, the feature maps in neural networks are splitted into local path and global path. Then, we define a LGD block to model the interaction between two paths as:
\begin{equation}\label{eq:block}
\begin{aligned}
\{ {\mathbf{x}}_{l} , {\mathbf{g}}_{l} \} = \mathcal{B} (\{ {\mathbf{x}}_{l-1} , {\mathbf{g}}_{l-1} \})
\end{aligned}~~,
\end{equation}
where $\{ {\mathbf{x}}_{l-1} , {\mathbf{g}}_{l-1} \}$ and $\{ {\mathbf{x}}_{l} , {\mathbf{g}}_{l} \}$ denote the input pair and output pair of the $l$-th block. The local-global pair consists of local feature map ${\mathbf{x}}_{l} \in {\mathbb{R}}^{C\times T\times H\times W}$ and global feature vector ${\mathbf{g}}_{l} \in {\mathbb{R}}^{C}$, where $C$, $T$, $H$ and $W$ are the number of channels, temporal length, height and width of the 4D volume data, respectively.

\begin{figure}[!tb]
   \centering {\includegraphics[width=0.4\textwidth]{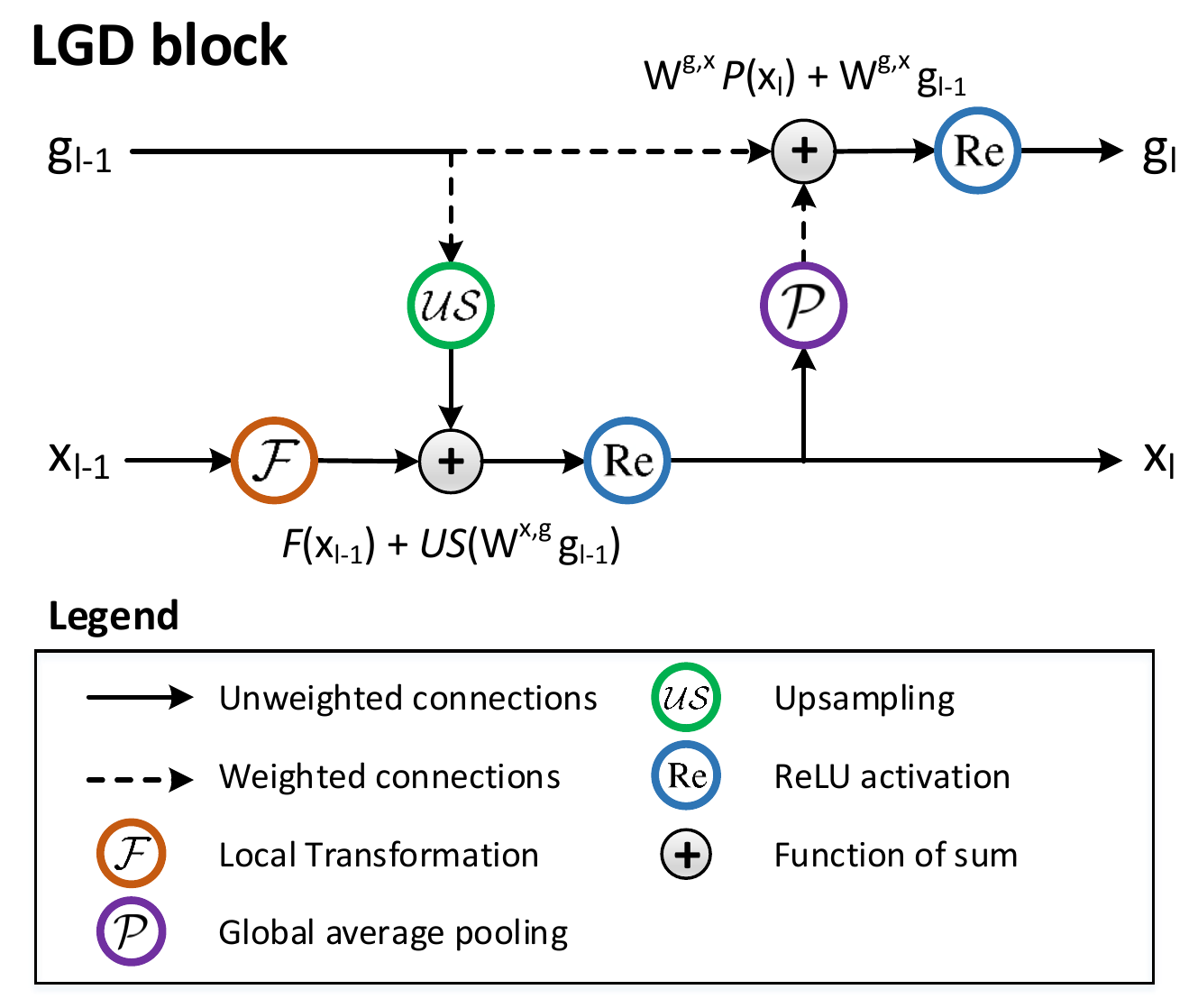}}
   \caption{\small A diagram of a LGD block. }
   \label{fig:block}
   \vspace{-0.15in}
\end{figure}

The detailed operations inside each block $\mathcal{B}$ are shown in Figure \ref{fig:block} and can be decomposed into two diffusion directions as following.

\textbf{(1) Global-to-local diffusion.} The first direction is to learn the transformation from ${\mathbf{x}}_{l-1}$ to the updated local feature ${\mathbf{x}}_{l}$ with the priority of global vector ${\mathbf{g}}_{l-1}$. Taking the inspiration from the recent successes of Residual Learning \cite{he2015deep}, we aim to formulate the global priority as the global residual value, which can be broadcasted to each location as
\begin{equation}\label{eq:local}
\begin{aligned}
{\mathbf{x}}_{l} = \text{ReLU} ( \mathcal{F} ({\mathbf{x}}_{l-1}) + \mathcal{US} (\mathbf{W}^{x,g} {\mathbf{g}}_{l-1}) )
\end{aligned}~~,
\end{equation}
where $\mathbf{W}^{x,g} \in {\mathbb{R}}^{C\times C}$ is the projection matrix, $\mathcal{US}$ is the upsampling operation duplicating the residual vector to each location and $\mathcal{F}$ is a local transformation function (i.e., 3D convolutions). The choice of function $\mathcal{F}$ is dependent on the network architecture and will be discussed in Section \ref{sec:LGDN}.

\textbf{(2) Local-to-global diffusion.} The second direction is to update the global vector with current local feature ${\mathbf{x}}_{l}$. Here, we simply linearly embed the input global feature ${\mathbf{g}}_{l-1}$ and Global Average Pooling (GAP) of local feature $\mathcal{P} ({\mathbf{x}}_{l})$ by
\begin{equation}\label{eq:global}
\begin{aligned}
{\mathbf{g}}_{l} = \text{ReLU} ( \mathbf{W}^{g,x} \mathcal{P} ({\mathbf{x}}_{l}) + \mathbf{W}^{g,g} {\mathbf{g}}_{l-1} )
\end{aligned}~~,
\end{equation}
where $\mathbf{W}^{g,x} \in {\mathbb{R}}^{C\times C}$ and $\mathbf{W}^{g,g} \in {\mathbb{R}}^{C\times C}$ are the projection matrices combining local and global features.

Compared with the traditional convolutional block which directly apply the transformation $\mathcal{F}$ to local feature, the LGD block introduced in Eq.(\ref{eq:local}) and Eq.(\ref{eq:global}) only requires three more projection matrices to produce the output pair. In order to reduce the additional parameters for LGD block, we exploit the low-rank approximation of each projection matrix as $\mathbf{W}=\mathbf{W}_{1} \mathbf{W}_{2}$, in which $\mathbf{W}_{1} \in {\mathbb{R}}^{C \times \hat{C}}$ and $\mathbf{W}_{2} \in {\mathbb{R}}^{\hat{C} \times C}$. When $\hat{C} \ll C$, the parameters as well as computational cost can be sharply reduced. Through cross-validation, we empirically set $\hat{C}=\frac{C}{16}$ which is found not to impact the performance negatively. By this approximation, the number of additional parameters is reduced from $3C^2$ to $\frac{3}{8}C^2$ for each block.

\begin{figure*}[!tb]
   \centering {\includegraphics[width=0.98\textwidth]{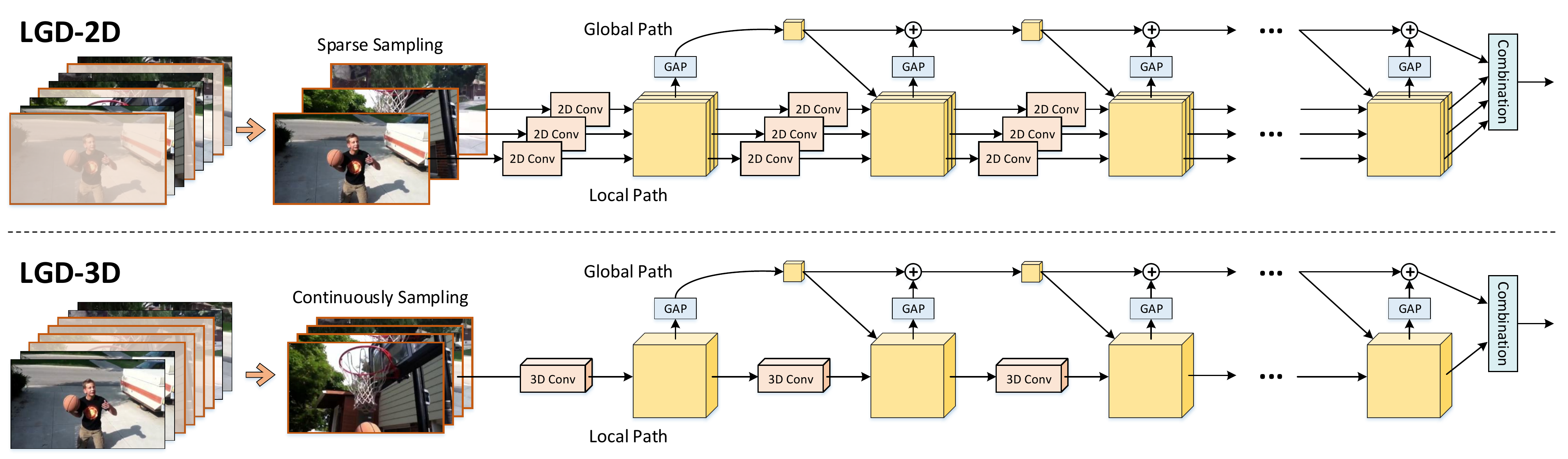}}
   \caption{\small The overview of two different Local and Global Diffusion networks. The upper one, called LGD-2D, applies the LGD block on the temporal segment network \cite{wang2016temporal}, which sparsely samples several frames and exploits 2D convolution as the local transformation. The lower one, called LGD-3D, continuously samples a short video clip and exploits pseudo-3D convolution \cite{qiu2017learning} as the local transformation. For both LGD networks, the learnt local and global features are combined to achieve the final representation.}
   \label{fig:framework}
   \vspace{-0.15in}
\end{figure*}

\subsection{Local and Global Combination Classifier} \label{fig:lgcc}
With the proposed LGD block, the network can learn local and global representations in parallel. The next question is how to make the final prediction by combining the two representations. Here, we consider the kernelized view of similarity measurement between two videos. Formally, denote $\{\mathbf{x}_{L}, \mathbf{g}_{L}\}$ and $\{\mathbf{x'}_{L}, \mathbf{g'}_{L}\}$ as the last output pair of two videos, we choose the bilinear kernel \cite{lin2015bilinear} on both the local and global features, which can be trained end-to-end in neural network. Thus, the kernel function can be given by
\begin{equation}\label{eq:bilinear}
\begin{aligned}
&k(\{\mathbf{x}_{L}, \mathbf{g}_{L}\}, \{\mathbf{x'}_{L}, \mathbf{g'}_{L}\})\\
&={\langle\mathbf{x}_{L}, \mathbf{x'}_{L} \rangle}_{2} + {\langle\mathbf{g}_{L}, \mathbf{g'}_{L} \rangle}_{2}\\
&=\frac{1}{N^2} \sum_{i} \sum_{j} {\langle\mathbf{x}^{i}_{L}, \mathbf{x'}^{j}_{L} \rangle}_{2} + {\langle\mathbf{g}_{L}, \mathbf{g'}_{L} \rangle}_{2}\\
&\approx\frac{1}{N^2} \sum_{i} \sum_{j} {\langle\varphi(\mathbf{x}^{i}_{L}), \varphi(\mathbf{x'}^{j}_{L}) \rangle} + {\langle\varphi(\mathbf{g}_{L}), \varphi(\mathbf{g'}_{L}) \rangle}
\end{aligned}~~,
\end{equation}
in which $N=L\times H \times W$ is the number of spatio-temporal locations, ${\langle \cdot, \cdot \rangle}_{2}$ is the bilinear kernel and $\mathbf{x}^{i}_{L} \in {\mathbb{R}}^{C}$ denotes the feature vector of $i$-th position in $\mathbf{x}_{L}$. In the last line of Eq (\ref{eq:bilinear}), we approximate the bilinear kernel by Tensor Sketch Projection $\varphi$ in \cite{gao2016compact}, which can effectively reduce the dimension of feature space.
By decomposing the kernel function in Eq (\ref{eq:bilinear}), the feature mapping is formulated as
\begin{equation}\label{eq:mapping}
\begin{aligned}
\phi(\{\mathbf{x}_{L}, \mathbf{g}_{L}\}) = [\frac{1}{N}\sum_{i}\varphi(\mathbf{x}^{i}_{L}), \varphi(\mathbf{g}_{L})]
\end{aligned}~~,
\end{equation}
where $[\cdot, \cdot ]$ denotes concatenation of two vectors. The $\phi(\{\mathbf{x}_{L}, \mathbf{g}_{L}\})$ combines the pair into a high dimensional representation. The whole process can be trained end-to-end in the neural networks. Finally, the resulting representation is fed into a fully connected layer for class labels prediction.

\section{Local and Global Diffusion Networks} \label{sec:LGDN}
The proposed LGD block and the classifier can be easily integrated with most of the existing video representation learning frameworks. Figure \ref{fig:framework} shows two different constructions of LGD blocks, called LGD-2D and LGD-3D, with different transformation $\mathcal{F}$ and training strategies.

\subsection{LGD-2D}
The straightforward way to learn video representation directly employs 2D convolution as the transformation function $\mathcal{F}$. Thus, in the local path of LGD-2D, a shared 2D CNN is performed as backbone network on each frame independently, as shown in the upper part in Figure \ref{fig:framework}. To enable efficient end-to-end learning, we uniformly split a video into T snippets and select only one frame per snippet for processing. The idea is inspired by Temporal Segment Network (TSN) \cite{wang2016temporal,wang2018temporal}, which overcomes computational issue by selecting a subset of frames for long-term temporal modeling. Thus, the input of LGD-2D consists of $T$ noncontinuous frames, and the global path learns a holistic representation of all these frames. Please note that the initial local representation $\mathbf{x}_{1}$ is achieved by a single local operation $\mathcal{F}$ applied on the input frames, and the initial global representation $\mathbf{g}_{1}=\mathcal{P}(\mathbf{x}_{1})$ is the global average of $\mathbf{x}_{1}$. At the end of the networks, the local and global combination classifier is employed to achieve a hybrid prediction.

\subsection{LGD-3D}
Another major branch of video representation learning is 3D CNN \cite{ji20133d,qiu2017learning,tran2015learning}. Following the common settings of 3D CNN, we feed $T$ consecutive frames into the LGD-3D network and exploit 3D convolution as local transformation $\mathcal{F}$, as shown in the lower part in Figure \ref{fig:framework}. Nevertheless, the training of 3D CNN is computationally expensive and the model size also has a quadratic growth compared with 2D CNN. Therefore, we choose the pseudo-3D convolution proposed in \cite{qiu2017learning} that decomposes 3D learning into 2D convolutions in spatial space and 1D operations in temporal dimension. To simplify the decomposition, in this paper, we only choose P3D-A block with the highest performance in \cite{qiu2017learning}, which cascades the the spatial convolution and temporal convolution in turn.

Here, we show the exampler architecture of LGD-3D based on the ResNet-50 \cite{he2015deep} backbone in Table \ref{tab:arch}. The LGD-3D firstly replaces each $3 \times 3$ convolutional kernel in original ResNet-50 with one $1 \times 3 \times 3$ spatial convolution and $3 \times 1 \times 1$ temporal convolution, and then builds a LGD block based on each residual unit. All the weights of spatial convolutions can be initialized from the pre-trained ResNet-50 model as done in \cite{qiu2017learning}. The dimension of input video clip is set as $16\times 112 \times 112$ consisting of $16$ consecutive frames with resolution $112 \times 112$. The clip length will be reduced twice by two max pooling layers with temporal stride of $2$. The computational cost and training time thus can be effectively reduced by the small input resolution and temporal pooling. The final local representation with dimension $4\times 7\times 7$ is combined with global representation by the kernelized classifier. This architecture can be easily extended to ResNet-101 or deeper networks by repeating more LGD blocks.

\subsection{Optimization} \label{sec:opt}

Next, we present the optimization of LGD networks. Considering the difficulty in training the whole network from scratch by kernelized classifier \cite{gao2016compact,lin2015bilinear}, we propose a two-stage strategy to train the LGD networks. At the beginning of the training, we optimize the basic network without the combination classifier, and adjust local and global representations separately. Denote $\{\mathbf{x}_{L}, \mathbf{g}_{L}\}$ and $\mathbf{y}$ as the last output pair and corresponding category of the input video, the optimization function is given as
\begin{equation}\label{eq:L1}
\begin{aligned}
\mathcal{L}_{\mathbf{W}_{g}}(\mathbf{g}_{L}, ~\mathbf{y}) + \mathcal{L}_{\mathbf{W}_{x}}(\mathcal{P}(\mathbf{x}_{L}), ~\mathbf{y})
\end{aligned}~~,
\end{equation}
where $\mathcal{L}_{\mathbf{W}}$ denotes the softmax cross-entropy loss with projection matrix $\mathbf{W}$. The overall loss consists of the classification errors from both global representation and local representation after global average pooling. After the training of basic network, we then tune the whole network with the following loss:
\begin{equation}\label{eq:L2}
\begin{aligned}
\mathcal{L}_{\mathbf{W}_{c}}(\phi(\{\mathbf{x}_{L}, \mathbf{g}_{L}\}), ~\mathbf{y})
\end{aligned}~~,
\end{equation}
where $\phi(\cdot)$ is the feature mapping proposed in Section \ref{fig:lgcc}.

\begin{table}
\centering
\scriptsize
\caption{\small The detailed architecture of LGD-3D with the ResNet-50 backbone network. The LGD blocks are shown in brackets and the kernel size for each convolution is presented followed by the number of output channels.}
\vspace{0.1cm}
\begin{tabular}{l|c|c} \hline
\begin{minipage}{0.88cm}\vspace{0.05cm} \textbf{Layer} \vspace{0.05cm}\end{minipage}     & \textbf{Operation} & \textbf{Local path size} \\ \hline
\begin{minipage}{0.88cm}\vspace{0.22cm} conv1 \vspace{0.22cm}\end{minipage}    & $\begin{matrix} 1\times7\times7, 64 \\ 3\times1\times1, 64 \end{matrix}$, stride 1, 2, 2 & $16\times56\times56$\\ \hline
\begin{minipage}{0.88cm}\vspace{0.10cm} pool1 \vspace{0.10cm}\end{minipage}    & $\begin{matrix} 2\times1\times1\end{matrix}$, max, stride 2, 1, 1 & $8\times56\times56$ \\ \hline
\begin{minipage}{0.88cm}\vspace{0.55cm} res2 \vspace{0.55cm}\end{minipage}    & ${\begin{bmatrix} 1\times1\times1,~64\\ 1\times3\times3,~64\\ 3\times1\times1,~64\\ 1\times1\times1,~256\end{bmatrix}}_{\text{LGD}}\times 3$ & $8\times56\times56$ \\ \hline
\begin{minipage}{0.88cm}\vspace{0.10cm} pool2 \vspace{0.10cm}\end{minipage}    & $\begin{matrix} 2\times1\times1\end{matrix}$, max, stride 2, 1, 1 & $4\times56\times56$ \\ \hline
\begin{minipage}{0.88cm}\vspace{0.55cm} res3 \vspace{0.55cm}\end{minipage}    & ${\begin{bmatrix} 1\times1\times1,~128\\ 1\times3\times3,~128\\ 3\times1\times1,~128\\ 1\times1\times1,~512\end{bmatrix}}_{\text{LGD}}\times 4$ & $4\times28\times28$ \\ \hline
\begin{minipage}{0.88cm}\vspace{0.55cm} res4 \vspace{0.55cm}\end{minipage}    & ${\begin{bmatrix} 1\times1\times1,~256\\ 1\times3\times3,~256\\ 3\times1\times1,~256\\ 1\times1\times1,~1024\end{bmatrix}}_{\text{LGD}}\times 6$ & $4\times14\times14$ \\ \hline
\begin{minipage}{0.88cm}\vspace{0.55cm} res5 \vspace{0.55cm}\end{minipage}    & ${\begin{bmatrix} 1\times1\times1,~512\\ 1\times3\times3,~512\\ 3\times1\times1,~512\\ 1\times1\times1,~2048\end{bmatrix}}_{\text{LGD}}\times 3$ & $4\times7\times7$ \\ \hline
\end{tabular}
\vspace{-0.15in}
\label{tab:arch}
\end{table}

\section{Experiments}
\subsection{Datasets}
We empirically evaluate LGD networks on the Kinetcis-400 \cite{carreira2017quo} and Kinetcis-600 \cite{ghanem2018activitynet} datasets. The Kinetics-400 dataset is one of the large-scale action recognition benchmarks. It consists of around 300K videos from 400 action categories. The 300K videos are divided into 240K, 20K, 40K for training, validation and test sets, respectively. Each video in this dataset is 10-second short clip cropped from the raw YouTube video. Note that the labels for test set are not publicly available and the performances on Kinetics-400 dataset are all reported on the validation set. The Kinetics-600 is an extended version of Kinetics-400 dataset, firstly made public in ActivityNet Challenge 2018 \cite{ghanem2018activitynet}. It consists of around 480K videos from 600 action categories. The 480K videos are divided into 390K, 30K, 60K for training, validation and test sets, respectively. Since the labels for Kinetics-600 test set are available, we report the final performance on both the validation and test sets.

\subsection{Training and Inference Strategy}
Our proposal is implemented on Caffe \cite{jia2014caffe} framework and the mini-batch Stochastic Gradient Descent (SGD) algorithm is employed to optimize the model. In the \textbf{training stage}, for LGD-2D, we set the input as $224 \times 224$ image which is randomly cropped from the resized $240 \times 320$ video frame. For LGD-3D, the dimension of input video clips is set as $16\times 112 \times 112$, which is randomly cropped from the resized non-overlapping 16-frame clip with the size of $16\times 120 \times 160$. Each frame/clip is randomly flipped along horizontal direction for data augmentation. We set each mini-batch as 128 triple frames for LGD-2D, and 64 clips for LGD-3D, which are implemented with multiple GPUs in parallel. The network parameters are optimized by standard SGD. For each stage in Section \ref{sec:opt}, the initial learning rate is set as 0.01, which is divided by 10 after every 20 epochs. The training is stopped after 50 epoches.

There are two \textbf{weights initialization} strategies for LGD networks. The first one is to train the whole networks from scratch. In this way, all the convolutional kernels and the projection matrices $\mathbf{W}$ in LGD block are initialized by Xavier initialization \cite{glorot2010understanding}, and all the biases are set as zero. The second one initializes the spatial convolutions with the existing 2D CNN pre-trained on ImageNet dataset \cite{russakovsky2015imagenet}. In order to keep the semantic information for these pre-trained convolutions, we set the projection matrix $\mathbf{W}^{x,g}$ as zero, making the global residual value vanishes when the training begins. Especially, the temporal convolutions in LGD-3D are initialized as an identity mapping in this case.

In the \textbf{inference stage}, we resize the video frames with the shorter side 240/120 for LGD-2D/LGD-3D, and perform spatially fully convolutional inference on the whole frame. Thus, the LGD-2D will predict one score for each triple frames and the video-level prediction score is calculated by averaging all scores from 10 uniformly sampled triple frames. Similarly, the video-level prediction score from LGD-3D is achieved by averaging all scores from 15 uniformly sampled 16-frame clips.

\subsection{Evaluation of LGD block}
\begin{table}
\centering
\caption{\small Performance comparisons between baseline and LGD block variants on Kinetics-600 validation set. All the backbone networks are ResNet-50 trained from scratch. The local and global combination classifier is not used for fair comparison.}
\begin{minipage}{0.23\textwidth}
\small
\centering
\vspace{0.1cm}
(a) LGD-2D
\vspace{0.1cm}
\end{minipage}
\begin{minipage}{0.23\textwidth}
\small
\centering
\vspace{0.1cm}
(b) LGD-3D
\vspace{0.1cm}
\end{minipage}
\begin{minipage}{0.23\textwidth}
\small
\centering
\begin{tabular}{@{~}l@{~}|@{~}c@{~}|@{~}c@{~}} \hline
\begin{minipage}{1.8cm}\vspace{0.12cm} \textbf{Method} \vspace{0.12cm}\end{minipage}     & \textbf{Top-1} & \textbf{Top-5} \\ \hline
\begin{minipage}{1.8cm}\vspace{0.12cm} TSN baseline \vspace{0.12cm}\end{minipage}     & 71.0 & 90.0 \\
\begin{minipage}{1.8cm}\vspace{0.12cm} block$_{v1}$ \vspace{0.12cm}\end{minipage}     & 71.6 & 90.2 \\
\begin{minipage}{1.8cm}\vspace{0.12cm} block$_{v2}$ \vspace{0.12cm}\end{minipage}     & 72.2 & 90.5 \\ \hline
\begin{minipage}{1.8cm}\vspace{0.12cm} \textbf{LGD block} \vspace{0.12cm}\end{minipage}     & \textbf{72.5} & \textbf{90.7} \\ \hline
\end{tabular}
\end{minipage}
\begin{minipage}{0.23\textwidth}
\small
\centering
\begin{tabular}{@{~}l@{~}|@{~}c@{~}|@{~}c@{~}} \hline
\begin{minipage}{1.8cm}\vspace{0.12cm} \textbf{Method} \vspace{0.12cm}\end{minipage}     & \textbf{Top-1} & \textbf{Top-5} \\ \hline
\begin{minipage}{1.8cm}\vspace{0.12cm} P3D baseline \vspace{0.12cm}\end{minipage}     & 71.2 & 90.5 \\
\begin{minipage}{1.8cm}\vspace{0.12cm} block$_{v1}$ \vspace{0.12cm}\end{minipage}     & 72.7 & 91.1 \\
\begin{minipage}{1.8cm}\vspace{0.12cm} block$_{v2}$ \vspace{0.12cm}\end{minipage}     & 73.6 & 91.6 \\ \hline
\begin{minipage}{1.8cm}\vspace{0.12cm} \textbf{LGD block} \vspace{0.12cm}\end{minipage}     & \textbf{74.2} & \textbf{92.0} \\ \hline
\end{tabular}
\end{minipage}
\vspace{-0.10in}
\label{tab:block}
\end{table}

\begin{figure}[!tb]
   \centering
   \subfigure[LGD-2D]{
     \label{fig:loss:a}
     \includegraphics[width=0.23\textwidth]{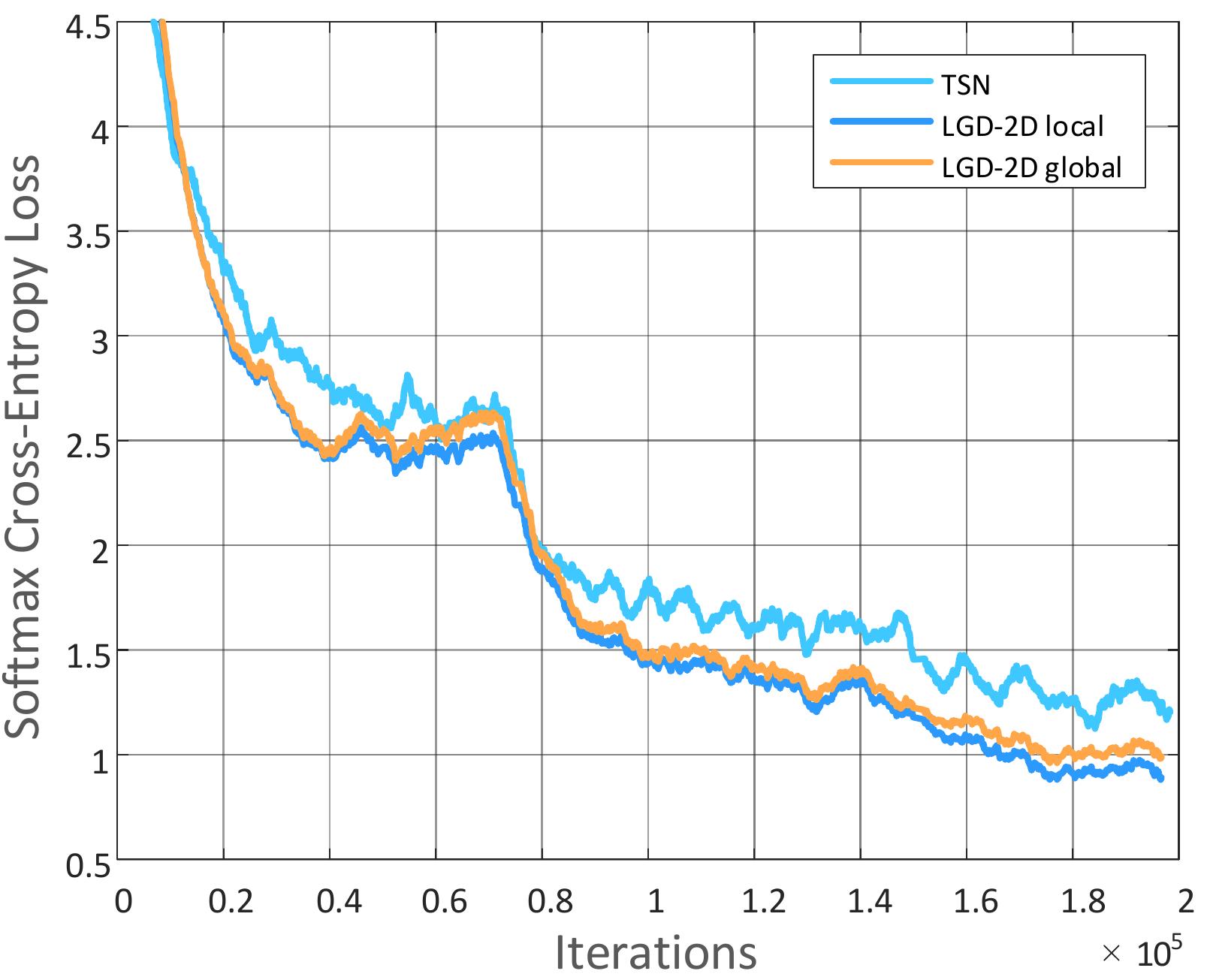}}
   \subfigure[LGD-3D]{
     \label{fig:loss:b}
     \includegraphics[width=0.23\textwidth]{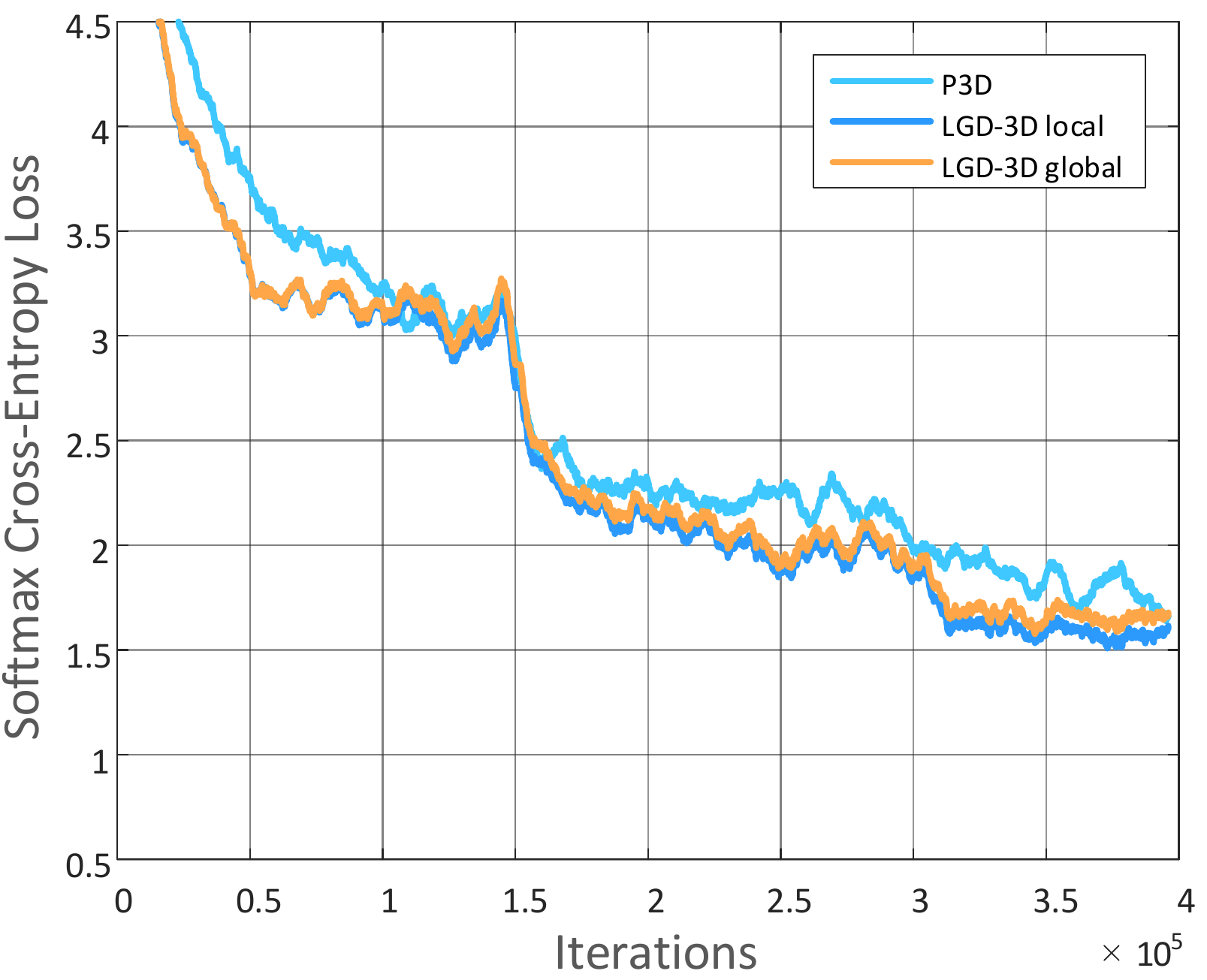}}
   \caption{\small The training loss on Kinetics-600 datasets. All the backbone networks in this figure are ResNet-50 trained from scratch.}
   \label{fig:loss}
   \vspace{-0.20in}
\end{figure}

We firstly verify the effectiveness of our proposed LGD block for spatio-temporal representation learning and compare with two diffusion block variants, i.e., block$_{v1}$ and block$_{v2}$ by different diffusion functions. Specifically, compared with LGD block, the \textbf{block$_{v1}$} ignores the global representation from lower layers, making the output function of global path as
\begin{equation}\label{eq:L1}
\begin{aligned}
{\mathbf{g}}_{l} = \mathcal{P} ({\mathbf{x}}_{l})
\end{aligned}~~.
\end{equation}
Motivated by the channel-wise scaling proposed in \cite{hu2018squeeze}, the \textbf{block$_{v2}$} utilizes the global priority as channel-wise multiplication. Thus, the output of local path in block$_{v2}$ can be formulated as
\begin{equation}\label{eq:L1}
\begin{aligned}
{\mathbf{x}}_{l} = \text{ReLU} ( \mathcal{F} ({\mathbf{x}}_{l-1}) \odot \mathcal{US} (sigmoid(\mathbf{W}^{x,g} {\mathbf{g}}_{l-1})) )
\end{aligned}~~,
\end{equation}
where $\odot$ denotes the element-wise multiplication.

Table \ref{tab:block} summarizes the performance comparisons on Kinetics-600 dataset. The backbone architectures are all ResNet-50 trained from scratch. Overall, all the three diffusion blocks (i.e., LGD block, block$_{v1}$ and block$_{v2}$) exhibit better performance than baseline networks for both 2D and 3D CNNs. The results basically indicate the advantage of exploring large-scale dependency by the diffusion between local path and global path. In particular, as indicated by our results, utilizing the proposed LGD block which embeds both input local and global representations and explores the global priority as residual value, can constantly lead to better performance than block$_{v1}$ and block$_{v2}$.

The loss curves of baseline networks and LGD networks are shown in Figure \ref{fig:loss}. The training losses of local and global paths in Eq. (\ref{eq:L1}) are given separately. Generally, the LGD networks produce lower losses than baseline networks, and converge faster and stably. Another observation is that the loss on local path is consistently lower than the loss on global path. We speculate that this may be due to information lost by low-rank approximation of projection matrices in Eq. (\ref{eq:global}).

\subsection{An Ablation Study of LGD networks}
\begin{table}
\centering
\small
\caption{\small Performance contribution of each design in LGD networks. Top-1 accuracies are shown on Kinetics-600 validation set.}
\vspace{0.1cm}
\begin{tabular}{l@{~}|ccccc|cc} \hline
\begin{minipage}{1.5cm}\vspace{0.12cm} \textbf{Method} \vspace{0.12cm}\end{minipage}     & \textbf{R50} & \textbf{R101} & \textbf{Img} & \textbf{Com} & \textbf{Long} & \textbf{Top-1} \\ \hline
\multirow{6}{*}{\textbf{LGD-2D}} &$\surd$&&&&& 72.5\\
&$\surd$&&$\surd$&&& 74.4\\
&$\surd$&&$\surd$&$\surd$&& 74.8\\ \cline{2-7}
&&$\surd$&&&& 74.5\\
&&$\surd$&$\surd$&&& 76.4\\
&&$\surd$&$\surd$&$\surd$&& 76.7\\ \hline
\multirow{8}{*}{\textbf{LGD-3D}} &$\surd$&&&&& 74.2\\
&$\surd$&&$\surd$&&& 75.8\\
&$\surd$&&$\surd$&$\surd$&& 76.3\\
&$\surd$&&$\surd$&$\surd$&$\surd$& 79.4\\ \cline{2-7}
&&$\surd$&&&& 76.0\\
&&$\surd$&$\surd$&&& 77.7\\
&&$\surd$&$\surd$&$\surd$&& 78.3\\
&&$\surd$&$\surd$&$\surd$&$\surd$& \textbf{81.5}\\ \hline
\end{tabular}
\vspace{-0.15in}
\label{tab:ablation}
\end{table}

Next, we study how each design in LGD networks influences the overall performance. Here, we choose ResNet-50 (\textbf{R50}) or ResNet-101 (\textbf{R101}) as backbone network. This backbone network is either trained from scratch or pre-trained by ImageNet (\textbf{Img}). The local and global combination classifier (\textbf{Com}) uses the kernelized classifier for prediction. In order to capture long-term temporal information, we further extend the LGD-3D network with 128-frame input (\textbf{Long}). Following the settings in \cite{wang2018non}, we firstly train the networks with 16-frame clips in the first stage in Section \ref{sec:opt} and then with 128-frame clips in the second stage. When training with 128-frame clips, we increase the stride of pool1 layer to 4, and set each mini-batch as 16 clips to meet the requirements of GPU memory. The training is stopped after 12.5 epoches.

Table \ref{tab:ablation} details the accuracy improvement on Kinetics-600 dataset by different designs of LGD networks. When exploiting ResNet-50 as backbone network, the pre-training on ImageNet dataset successfully boosts up the top-1 accuracy from 72.5\% to 74.4\% for LGD-2D and from 74.2\% to 75.8\% for LGD-3D. This demonstrates the effectiveness of pre-training on large-scale image recognition dataset. The local and global combination classifier which combines the representations from two paths leads to the performance boost of 0.4\% and 0.5\% for LGD-2D and LGD-3D, respectively. Especially for LGD-3D, the training on 128-frame clips contributes a large performance increase of 3.1\% by involving long-term temporal information in the network. Moreover, compared with ResNet-50, both the LGD-2D and LGD-3D based on ResNet-101 exhibit significantly better performance, with the top-1 accuracy of 76.7\% and 81.5\% for LGD-2D and LGD-3D, respectively. The results verify that deeper networks have larger learning capacity for spatio-temporal representation learning.

\subsection{Comparisons with State-of-the-Art}
\begin{table}
\centering
\small
\caption{\small Performance comparisons with the state-of-the-art methods on Kinetics-400 validation set. }
\vspace{0.1cm}
\begin{tabular}{@{~}l@{~}c@{~}|c|c@{~~}} \hline
\begin{minipage}{1.5cm}\vspace{0.12cm} \textbf{Method} \vspace{0.12cm}\end{minipage} & \textbf{Backbone} & \textbf{Top-1} & \textbf{Top-5} \\ \hline
I3D RGB \cite{carreira2017quo} & Inception & 72.1 & 90.3 \\
I3D Flow \cite{carreira2017quo} & Inception & 65.3 & 86.2 \\
I3D Two-stream \cite{carreira2017quo} & Inception & 75.7 & 92.0 \\
ResNeXt-101 RGB \cite{hara2018can} & custom & 65.1 & 85.7 \\
R(2+1)D RGB \cite{tran2018closer} & custom & 74.3& 91.4\\
R(2+1)D Flow \cite{tran2018closer} & custom & 68.5& 88.1\\
R(2+1)D Two-stream \cite{tran2018closer} & custom & 75.4& 91.9\\
NL I3D RGB \cite{wang2018non} & ResNet-101 & 77.7 & 93.3 \\
S3D-G RGB \cite{xie2018rethinking} & Inception & 74.7 & 93.4\\
S3D-G Flow \cite{xie2018rethinking} & Inception & 68.0 & 87.6  \\
S3D-G Two-stream \cite{xie2018rethinking} & Inception & 77.2 & 93.0 \\\hline
\multicolumn{2}{l|}{From Anet17 winner report \cite{bian2017revisiting}} &&\\
2D CNN RGB & Inception-ResNet-v2  & 73.0 & 90.9 \\
Three-stream late fusion & Inception-ResNet-v2 & 74.9 & 91.6 \\
Three-stream SATT & Inception-ResNet-v2 & 77.7 & 93.2 \\ \hline
\textbf{LGD-3D RGB} & ResNet-101 & 79.4 & 94.4 \\
\textbf{LGD-3D Flow} & ResNet-101 & 72.3 & 90.9 \\
\textbf{LGD-3D Two-stream} & ResNet-101 & \textbf{81.2} & \textbf{95.2} \\ \hline
\end{tabular}
\vspace{-0.15in}
\label{tab:k400}
\end{table}

We compare with several state-of-the-art techniques on Kinetics-400 and Kinetics-600 datasets. The performance comparisons are summarized in tables \ref{tab:k400} and \ref{tab:k600}, respectively. Please note that most recent works employ fusion of two or three modalities on these two datasets. Broadly, we can categorize the most common modalities into four categories, i.e., RGB, Flow, Two-stream and Three-stream. The \textbf{RGB}/\textbf{Flow} feeds the video frames/optical flow images into the networks. The optical flow image in this paper consists of two-direction optical flow extracted by TV-L1 algorithm \cite{zach2007duality}. The predictions from RGB and Flow modalities are fused by \textbf{Two-stream} methods. The \textbf{Three-stream} approaches further merge the prediction from audio input.

As shown in Table \ref{tab:k400}, with only RGB input, the LGD-3D achieves 79.4\% top-1 accuracy, which makes the relative improvement over the recent approaches I3D \cite{carreira2017quo}, R(2+1)D \cite{tran2018closer}, NL I3D \cite{wang2018non} and S3D-G \cite{xie2018rethinking} by 10.1\%, 6.8\%, 2.1\% and 6.2\%, respectively. This accuracy is also higher than 2D CNN with a deeper backbone reported by the ActivityNet 2017 challenge winner \cite{bian2017revisiting}. Note that the LGD-3D with RGB input can obtain higher performance even compared with the Two-stream or Three-stream methods. When fusing the prediction from both RGB and Flow modalities, the accuracy of LGD-3D will be further improved to 81.2\%, which is to-date the best published performance on Kinetics-400.

\begin{table}
\centering
\small
\caption{\small Performance comparisons with the state-of-the-art methods on Kinetics-600. Most of the performances are reported on validation set except the performance of LGD-3D Two-stream* are on the test set.}
\vspace{0.1cm}
\begin{tabular}{@{~~}l@{~}c@{~}|c|c@{~~}} \hline
\begin{minipage}{1.5cm}\vspace{0.12cm} \textbf{Method} \vspace{0.12cm}\end{minipage} & \textbf{Backbone} & \textbf{Top-1} & \textbf{Top-5} \\ \hline
\multicolumn{2}{l|}{From Anet18 winner report \cite{he2018exploiting}} &&\\
TSN RGB & SENet-152 & 76.2 & -- \\
TSN Flow & SENet-152 & 71.3 & -- \\
StNet RGB & Inception-ResNet-v2  & 78.9 & -- \\
NL I3D RGB & ResNet-101 & 78.6 & -- \\
Three-stream Attention & mixed & 82.3 & 96.0 \\
Three-stream iTXN & mixed & 82.4 & 95.8 \\ \hline
\multicolumn{2}{l|}{From Anet18 runner-up report \cite{yao2018yh}} &&\\
P3D RGB & ResNet-152 & 78.4 & 93.9 \\
P3D Flow & ResNet-152 & 71.0 & 90.0 \\
P3D Two-stream & ResNet-152 & 80.9 & 94.9 \\ \hline
\textbf{LGD-3D RGB} & ResNet-101 & 81.5 & 95.6 \\
\textbf{LGD-3D Flow} & ResNet-101 & 75.0 & 92.4 \\
\textbf{LGD-3D Two-stream} & ResNet-101 & \textbf{83.1} & \textbf{96.2} \\ \hline
\textbf{LGD-3D Two-stream*} & ResNet-101 & \textbf{82.7} & \textbf{96.0} \\ \hline
\end{tabular}
\vspace{-0.15in}
\label{tab:k600}
\end{table}

Similar results are also observed on Kinetics-600, as summarized in Table \ref{tab:k600}. Since this dataset is recently made available for ActivityNet 2018 challenge, we show the performance of different approaches reported by the challenge winner \cite{he2018exploiting} and challenge runner-up \cite{yao2018yh}. With the RGB inputs, LGD-3D achieves 81.5\% top-1 accuracy on Kinetics-600 validation set, which obtains 3.4\% relative improvement than P3D with the deeper backbone of ResNet-152. The performance is higher than that of NL I3D which also explores large-range dependency. This result basically indicates that LGD network is an effective way to learn video representation with a global aspect. By combining the RGB and Flow modalities, the top-1 accuracy of LGD-3D achieves 83.1\%, which is even higher than three-stream method proposed by ActivityNet 2018 challenge winner.

\subsection{Evaluation on Video Representation}
Here we evaluate video representation learnt by our LGD-3D for two different tasks and on four popular datasets, i.e., UCF101, HMDB51, J-HMDB and UCF101D. UCF101 \cite{UCF101} and HMDB51\cite{HMDB51} are two of the most popular video action recognition benchmarks. UCF101 consists of 13K videos from 101 action categories, and HMDB51 consists of 7K videos from 51 action categories. We follow the three training/test splits provided by the dataset organisers. Each split in UCF101 includes about 9.5K training and 3.7K test videos, while a HMDB51 split contains 3.5K training and 1.5K test videos.

J-HMDB and UCF101D are two datasets for spatio-temporal action detection. J-HMDB \cite{jhuang2013towards} contains 928 well trimmed video clips of 21 actions. The videos are truncated to actions and the bounding box annotations are available for all frames. It provides three training/test splits for evaluation. UCF101D \cite{UCF101} is a subset of UCF101 for action detection task. It consists of 3K videos from 24 classes with spatio-temporal ground truths.

\begin{table}
\centering
\small
\caption{\small Performance comparisons with the state-of-the-art methods on UCF101 (3 splits) and HMDB51 (3 splits).}
\vspace{0.1cm}
\begin{tabular}{l@{~}c@{~}|@{~}c@{~}|@{~}c@{~~}} \hline
\begin{minipage}{1.5cm}\vspace{0.12cm} \textbf{Method} \vspace{0.12cm}\end{minipage} & \textbf{Pretraining} & \textbf{U101} & \textbf{H51} \\ \hline
IDT \cite{wang2013action} & -- & 86.4 & 61.7 \\
Two-stream \cite{simonyan2014two} & ImageNet & 88.0 & 59.4 \\
TSN \cite{wang2016temporal} & ImageNet & 94.2 & 69.4 \\ \hline
I3D RGB \cite{carreira2017quo} & ImageNet+Kinetics-400 & 95.4 & 74.5 \\
I3D Flow \cite{carreira2017quo} & ImageNet+Kinetics-400 & 95.4 & 74.6\\
I3D Two-stream \cite{carreira2017quo} & ImageNet+Kinetics-400 & 97.9 & 80.2 \\
ResNeXt-101 RGB \cite{hara2018can} & Kinetics-400 & 94.5 & 70.2 \\
R(2+1)D RGB \cite{tran2018closer} & Kinetics-400 & 96.8& 74.5\\
R(2+1)D Flow \cite{tran2018closer} & Kinetics-400 & 95.5& 76.4\\
R(2+1)D Two-stream \cite{tran2018closer} & Kinetics-400 & 97.3& 78.7\\
S3D-G RGB \cite{xie2018rethinking} & ImageNet+Kinetics-400 & 96.8 & 75.9\\ \hline
\textbf{LGD-3D RGB} & ImageNet+Kinetics-600 & 97.0 & 75.7 \\
\textbf{LGD-3D Flow} & ImageNet+Kinetics-600 & 96.8 & 78.9 \\
\textbf{LGD-3D Two-stream} & ImageNet+Kinetics-600 & \textbf{98.2} & \textbf{80.5} \\ \hline
\end{tabular}
\vspace{-0.05in}
\label{tab:u101h51}
\end{table}

\begin{table}
\centering
\small
\caption{\small The performance in terms of video-mAP on J-HMDB (3 splits) and UCF101D datasets.}
\vspace{0.1cm}
\begin{tabular}{l@{~}|c@{~~~}c|c@{~~~}c@{~~~}c@{~~~}c} \hline
\multirow{2}{*}{\textbf{Method}}     & \multicolumn{2}{c|}{\textbf{J-HMDB}} & \multicolumn{4}{c}{\textbf{UCF101D}} \\ \cline{2-7}
                                     & 0.2 & 0.5 & 0.05 & 0.1 & 0.2 & 0.3 \\ \hline
\begin{minipage}{3cm}\vspace{0.08cm} Weinzaepfel \emph{et al.} \cite{weinzaepfel2015learning} \vspace{0.08cm}\end{minipage} & 63.1 & 60.7 & 54.3 & 51.7 & 46.8 & 37.8 \\
\begin{minipage}{3cm}\vspace{0.08cm} Saha \emph{et al.} \cite{saha2016deep} \vspace{0.08cm}\end{minipage} & 72.6 & 71.5 & 79.1 & 76.6 & 66.8 & 55.5 \\
\begin{minipage}{3cm}\vspace{0.08cm} Peng \emph{et al.} \cite{peng2016multi} \vspace{0.08cm}\end{minipage} & 74.3 & 73.1 & 78.8 & 77.3 & 72.9 & 65.7 \\
\begin{minipage}{3cm}\vspace{0.08cm} Singh \emph{et al.} \cite{singh2017online} \vspace{0.08cm}\end{minipage} & 73.8 & 72.0 & -- & -- & 73.5 & -- \\
\begin{minipage}{3cm}\vspace{0.08cm} Kalogeiton \emph{et al.} \cite{kalogeiton2017action} \vspace{0.08cm}\end{minipage} & 74.2 & 73.7 & -- & -- & 77.2 & -- \\
\begin{minipage}{3cm}\vspace{0.08cm} Hou \emph{et al.} \cite{hou2017tube} \vspace{0.08cm}\end{minipage} & 78.4 & 76.9 & 78.2 & 77.9 & 73.1 & 69.4 \\
\begin{minipage}{3cm}\vspace{0.08cm} Yang \emph{et al.} \cite{yang2017spatio} \vspace{0.08cm}\end{minipage} & -- & -- & 79.0 & 77.3 & 73.5 & 60.8 \\
\begin{minipage}{3cm}\vspace{0.08cm} Li \emph{et al.} \cite{li2018recurrent} \vspace{0.08cm}\end{minipage} & 82.7 & 81.3 & 82.1 & 81.3 & 77.9 & 71.4 \\ \hline
\begin{minipage}{3cm}\vspace{0.08cm} \textbf{LGD-3D RGB} \vspace{0.08cm}\end{minipage} & 77.3 & 74.2 & 78.8 & 77.6 & 69.3 & 64.1 \\
\begin{minipage}{3cm}\vspace{0.08cm} \textbf{LGD-3D Flow} \vspace{0.08cm}\end{minipage} &  84.5 & 82.9 & 86.5 & 84.2 & 79.8 & 74.7 \\
\begin{minipage}{3cm}\vspace{0.08cm} \textbf{LGD-3D Two-stream} \vspace{0.08cm}\end{minipage} & \textbf{85.7} & \textbf{84.9} & \textbf{88.3} & \textbf{87.1} & \textbf{82.2} & \textbf{75.6} \\ \hline
\end{tabular}
\vspace{-0.15in}
\label{tab:detection}
\end{table}

We first validate the global representations learnt by the pre-trained LGD-3D network. Therefore, we fine-tune the pre-trained LGD-3D on the UCF101 and HMDB51 datasets. The performance comparisons are summarized in Table \ref{tab:u101h51}. Overall, the two-stream LGD-3D achieves 98.2\% on UCF101 and 80.5\% on HMDB51, which consistently indicate that video representation produced by our LGD-3D attains a performance boost against baselines on action recognition task. Specifically, the two-stream LGD-3D outperforms three traditional approaches, i.e., IDT, Two-stream and TSN by 11.8\%, 10.2\% and 4.0\% on UCF101, respectively. The results demonstrate the advantage of pre-training on large-scale video recognition dataset. Moreover, compared with recent methods pre-trained on Kinetics-400 dataset, LGD-3D still surpasses the best competitor Two-stream I3D by 0.3\% on UCF101.

\begin{figure}[!tb]
   \centering {\includegraphics[width=0.49\textwidth]{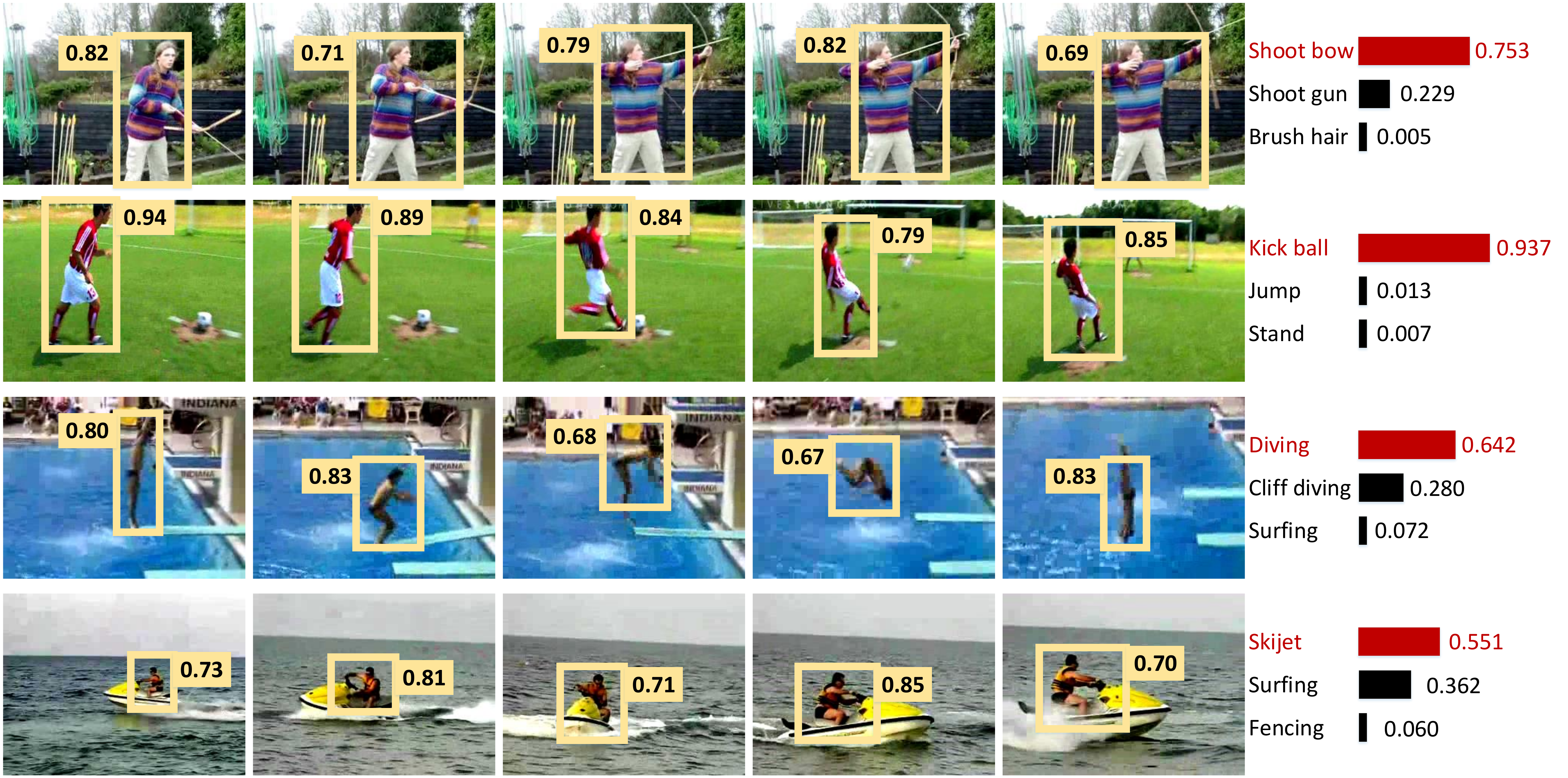}}
   \caption{\small Four detection examples of our method from J-HMDB (upper two rows) and UCF101D (lower two rows). The proposal score is given for each bounding box. Top predicted action classes for each tubelet are on the right.}
   \label{fig:Examples}
   \vspace{-0.15in}
\end{figure}

Next, we turn to evaluate the local representations from pre-trained LGD-3D networks on the task of spatio-temporal action detection. To build the action detection framework based on LGD-3D, we firstly obtain the action proposals in each frame by a region proposal network \cite{ren2015faster} with ResNet-101. The action tubelet is generated by proposal linking and temporally trimming in \cite{saha2016deep}. Then the prediction score of each proposal is estimated by the ROI-pooled local feature from LGD-3D network. In Table \ref{tab:detection}, we summarize the performance comparisons on J-HMDB (3 splits) and UCF101D with different IoU thresholds. Our LGD-3D achieves the best performance at all the cases. Specifically, at the standard threshold (0.5 for J-HMDB, and 0.2 for UCF101D), LGD-3D makes relative improvement of 4.4\% and 5.5\% than the best competitor \cite{li2018recurrent} on J-HMDB and UCF101D, respectively. Figure \ref{fig:Examples} showcases four detection examples from J-HMDB and UCF101D.

\section{Conclusion}
We have presented Local and Global Diffusion (LGD) network architecture which aims to learn local and global representations in an unified fashion. Particularly, we investigate the interaction between localized and holistic representations, by designing LGD block with diffusion operations to model local and global features.
A kernelized classifier is also formulated to combine the final prediction from two representations. With the development of the two components, we have proposed two LGD network architectures, i.e., LGD-2D and LGD-3D, based on 2D CNN and 3D CNN, respectively. The results on large-scale Kinetics-400 and Kinetics-600 datasets validate our proposal and analysis. Similar conclusion is also drawn from the other four datasets in the context of video action recognition and spatio-temporal action detection. The spatio-temporal video representation produced by our LGD networks is not only effective but also highly generalized across datasets and tasks.
Performance improvements are clearly observed when comparing to other feature learning techniques. More remarkably, we achieve new state-of-the-art performances on all the six datasets.

Our future works are as follows. First, more advanced techniques, such as attention mechanism, will be investigated in the LGD block. Second, more in-depth study of how to combine the local and global representations could be explored. Third, we will extend the LGD network to other types of inputs, e.g., audio information.

{\small
\bibliographystyle{ieee_fullname}
\bibliography{egbib}
}

\end{document}